\documentclass[letterpaper, 10 pt, conference]{ieeeconf} 

\IEEEoverridecommandlockouts
\usepackage{cite}
\usepackage{amsmath,amssymb,amsfonts}
\usepackage{algorithmic}
\usepackage{graphicx}
\usepackage{textcomp}
\usepackage{xcolor}
\usepackage{balance}
\usepackage[hidelinks]{hyperref}
\usepackage{lipsum}
\usepackage{breqn}
\usepackage{booktabs}
\usepackage{siunitx}
\usepackage{adjustbox}
\usepackage{tikz}
\usetikzlibrary{calc,patterns,decorations.pathmorphing,decorations.markings,positioning,backgrounds,arrows.meta,shapes,fit,matrix,spy,shapes.geometric}

\begin{document}


\title{\LARGE \bf A Passive Variable Impedance Control Strategy with Viscoelastic Parameters Estimation of Soft Tissues for Safe Ultrasonography
\thanks{We acknowledge the support of the MUR PNRR project FAIR - Future AI Research (PE00000013) and the project iNEST - Interconnected Nord-Est Innovation Ecosystem (ECS 00000043) funded by the NextGenerationEU.}}

\author{Luca Beber$^{1}$, Edoardo Lamon$^{2,3}$, Davide Nardi$^{2}$, Daniele Fontanelli$^{1}$, Matteo Saveriano$^{1}$, Luigi Palopoli$^{2}$
\thanks{$^{1}$Department of Industrial Engineering, Universit\`a di Trento, Trento, Italy.}
\thanks{$^{2}$Department of Information Engineering and Computer Science, Universit\`a di Trento, Trento, Italy. \tt\small edoardo.lamon@unitn.it}
\thanks{$^{3}$Human-Robot Interfaces and Interaction, Istituto Italiano di Tecnologia, Genoa, Italy.}
}

\maketitle

\thispagestyle{empty}
\pagestyle{empty}

\begin{abstract}
In the context of telehealth, robotic approaches have proven a valuable solution to in-person visits in remote areas, with decreased costs for patients and infection risks. In particular, in ultrasonography, robots have the potential to reproduce the skills required to acquire high-quality images while reducing the sonographer's physical efforts. In this paper, we address the control of the interaction of the probe with the patient's body, a critical aspect of ensuring safe and effective ultrasonography. We introduce a novel approach based on variable impedance control, allowing the real-time optimisation of compliant controller parameters during ultrasound procedures. This optimisation is formulated as a quadratic programming problem and incorporates physical constraints derived from viscoelastic parameter estimations. Safety and passivity constraints, including an energy tank, are also integrated to minimise potential risks during human-robot interaction. The proposed method's efficacy is demonstrated through experiments on a patient's dummy torso, highlighting its potential for achieving safe behaviour and accurate force control during ultrasound procedures, even in cases of contact loss. 

\end{abstract}

\section{Introduction}
Telehealth is commonly defined as the ability to provide health care services, removing the need for in-person visits. This paradigm is particularly useful to reach geographically remote areas and has gained traction during the Covid-19 pandemics for the obvious reduction in the risk of pathogen transmission between patients and medical staff. The recognised advantages of this approach are increased comfort for patients and cost reduction for both patients and the health service~\cite{eu2018market,who2022consolidated}.
The most advanced frontier of telehealth is the ability for the physician to execute complex diagnostic activities that require physical contact with the patient. An example of this kind is tactile examination (palpation), in which the doctor uses fingertips and palms to feel the presence of anomalies underneath the skin. A promising solution for this type of activity is telerobotics~\cite{avgousti2016medical}, in which a robotic system interacts with the patient and is remotely controlled or supervised by a physician. Medical applications of telerobotics are perceived as technically challenging because they require the integration of different tools, e.g., virtual reality, haptic interfaces, precise control, and force feedback. In the rich set of possible applications of medical telerobotics, ultrasonography stands out as one of the most promising.
It is a safe, noninvasive diagnostic tool that allows healthcare professionals to evaluate the state of organs and issues in order to detect the potential presence of diseases or assess their evolution. Acquiring ultrasound images is a complex task that requires skilled sonographers and the continuous exertion of considerable force, which might result in a health risk (work-related musculoskeletal disorders\cite{coffin2014work}).  The quality of the diagnosis very much depends on the skills of the sonographer, and the number of experienced operators is not sufficient to deliver the service in remote areas~\cite{li2021overview}. Robotised solutions, which involve lightweight manipulators and accurate perception systems, have the potential to eliminate such issues. 
Key enablers are a new generation of sensorised probes~\cite{ma2022asee} and path planning strategies computed from RGB-D cameras~\cite{tan2022automatic}, which can adapt to the motion of the patient's tissues through online path updates~\cite{hennersperger2017towards,zhan2020autonomous}.

A crucial aspect is how to control the motion of the probe along trajectories on the patient's body. The literature in this area offers both autonomous~\cite{li2021overview,roshan2022robotic} and teleoperated~\cite{vonhaxthausen2021medical,jiang2023robotic} solutions. While the two approaches differ mainly on position reference computation, both require the regulation of the contact forces with the patient's body.
Indeed, applying excessive force during this contact can potentially distort the target anatomical structure and cause harm to the patient. Conversely, insufficient force will not guarantee effective transmission of acoustic waves, leading to poor image quality. Therefore, the goal of most existing robot control methods for ultrasonography is to apply a constant force on the probe in the normal direction of the patient's surface~\cite{merouche2016robotic, hennersperger2017towards, tsumura2020robotic}. 
Tsumura et al. \cite{tsumura2020robotic} advocate the use of passive spring as an alternative to the robot's actuators in order to guarantee by design that the maximum force will be below an acceptable limit, preserving the patient's safety. 
\textit{Virga et al.}~\cite{virga2016automatic} proposed to apply a patient-specific optimal contact force encoded through the ultrasound confidence map, which contains prior information on the ultrasound image quality. Other approaches, instead, tailor the force to the patient by means of the estimates of biomechanics properties of the patient's tissues~\cite{pappalardo2016hunt,ferro2021online}. Nevertheless, through force control, the direction tangential to the body surface cannot be position-controlled; therefore, a correct operation of the system  requires to switch between position (free space motions) and force control (when in contact with the body). Another effective way to indirectly control the probe force is through task-space compliant schemes. In \cite{jiang2021automatic}, for instance, the robot desired force consists of a constant term plus a spring with constant stiffness. \textit{Wang et al.} instead model the impedance as a mass-spring-damper system to control a robot through a hybrid admittance strategy with constant parameters~\cite{wang2023task}. These approaches appear promising when the tissue or organ to be scanned has approximately the same biomechanics properties and the parameters can be tuned accordingly. However, some specific exams, such as lung or heart ultrasound, require the probe to pass through bones and soft parts in close proximity (e.g., chest and abdomen). This scenario reveals the requirement for adapting compliance to different situations. For example, \textit{Duan et al.} propose to transfer human motion skills to the robot by learning the impedance from human demonstrations~\cite{duan2022ultrasound}. 
An example of this is when the patient moves or there is a contact loss between the probe tip and the patient's body (due also to the presence of a gel that removes the friction of the surface).  
In this scenario, monitoring and restricting the energy and power transmission is of the greatest importance in achieving a safe human-robot interaction. 

In this paper, we focus on the problem of force-controlled motion for lung and heart ultrasonography. At the core of our approach is a method to optimise on-the-fly the impedance parameters of a compliant controller by exploiting the paradigm of variable impedance control. The optimisation problem is formulated through quadratic programming (QP) and includes physical constraints, which are obtained by means of a prior estimation of the viscoelastic parameters, and safety constraints through the addition of an energy tank. To initialise the proposed control strategy, an offline phase is required, which consists of a discrete biomechanics characterisation and a smoothing operation to retrieve a continuous body description. 
The nonlinear model proposed by \textit{Hunt and Crossley}~\cite{hunt1975coefficient} is used to relate the deformation to the tissue force, and the viscoelastic parameters are retrieved by minimising the error with respect to the forces measured by a F/T sensor. Then, the estimated parameters are interpolated through a Gaussian Process (GP) to describe the desired body surface smoothly.  In the online phase, these parameters are used in the computation of impedance parameters to ensure that the correct amount of force is tracked and that the penetration into the body is limited, removing the requirement for precise control in cases of perception inaccuracy or failure. 
In addition, an energy tank is used to limit both the energy introduced into the system and its power flow to prevent a large amount of energy from being injected instantaneously. The method is evaluated in a proof-of-concept ultrasound of the dummy torso of a patient through a set of experiments that evaluate the viscoelastic modelling and compare the performance of the proposed passive variable impedance with standard approaches. 

\section{Methodology}
\subsection{Tissue Parameters Estimation}\label{ssec:parameters_estimation}

Biological tissues are known to demonstrate viscoelastic behaviour, implying that their response depends not only on the deformation applied but also on the rate of deformation. As a result, they can be represented using springs and dampers arranged in various configurations~\cite{ozkaya2012fundamentals}. The simplest and most common model is the Kelvin-Voight model, where the tissue is modelled with a spring damper-system~\cite{flugge1975viscoelasticity}. 
However, this model has major limitations especially when the
penetration is small. As shown
in~\cite{diolaiti2005contact,hunt1975coefficient}, the hysteresis loop
obtained with the Kelvin-Voight model is not energetically
consistent. Moreover, the contact phase is characterised by unnatural
behaviours since, using this model, the coefficient of restitution is
not dependent on the velocity~\cite{diolaiti2005contact}. To overcome
these limitations, non-linear models, where the velocity is also
dependent on the penetration depth, have been developed.  One of the
most used is the Hunt-Crossley (HC) model~\cite{hunt1975coefficient}
\begin{equation}
F_{tissue}(t)= \begin{cases}\kappa \varepsilon^{\beta}(t)+\lambda \varepsilon^{\beta}(t) \dot{\varepsilon}(t), & \varepsilon \geq 0 \\ 0, & \varepsilon<0\end{cases},
\label{eq:hunt_crossley}
\end{equation}
where $\varepsilon(t)\in\mathbb{R}$ is the amount of penetration, $\dot{\varepsilon}(t)\in\mathbb{R}$ is the penetration rate, $\kappa\in\mathbb{R}$ is the elasticity, and $\eta\in\mathbb{R}$ is the viscosity. Note that these coefficients do not correspond to the real values of elasticity and viscosity, but for ease of communication, we will refer to them as such. The exponent $\beta$ takes into account the variation of the contact area between the indenter and the surface as the penetration depth changes. $\beta$ depends on both the type of material and the shape of the tip being used. 
Values $\beta \in (1.0, 1.5)$ are generally better for materials that are quite soft, as in the case of biological tissue. The hysteresis loop obtained with the Hunt-Crossley model is energetically consistent (i.e., the loop closes). Moreover, the restitution factor depends on the velocity of impact but it is independent from the exponent $\beta$~\cite{diolaiti2005contact}, resulting in a more natural contact behaviour. 

Due to the presence of viscous characteristics a dynamic test is
necessary to identify the parameters
of~\eqref{eq:hunt_crossley}. Therefore, we control the robot to
perform a sinusoidal motion in the vertical direction and collect data
(force and penetration) for the estimation process. Considering that
the robot's end-effector always remains in contact with the surface
during the data collection, the force exerted by the tissue is
\begin{equation}
  F_{tissue} = \kappa ( s - z_{ee})^{\beta} - \lambda \dot{z}_{ee} ( s - z_{ee})^{\beta},
  \label{eq:hunt_cross_ee}
\end{equation}
where $s$ and $z_{ee}$ are the position of the tissue surface and the
position of the end-effector along the $z$-axis. The minus sign
indicates that the penetration rate is in the opposite direction of
the end-effector velocity. The force collected at the force sensor is
\begin{align}
    F_{sensor} &= F_{tissue} - m_{I} \Ddot{z}_{ee}\\
    &= \kappa ( s - z_{ee})^{\beta} - \lambda \dot{z}_{ee} ( s - z_{ee})^{\beta} - m_{I} \Ddot{z}_{ee},
\end{align}
where $m_I$ is the mass of the indenter. Parameters estimation are derived offline using a least square algorithm to find
the parameters $\kappa$ and $\lambda$ that best fit the sensed force
profile. The least square minimises the sum-of-square loss
\begin{dmath}
    \mathcal{L} = \sum_{j=i}^{n} \left((F_{sensor,j} + m_I \Ddot{z}_{ee,j}) - ( \kappa ( s_j - z_{ee,j})^{\beta} - \lambda \dot{z}_{ee,j} ( s_j - z_{ee,j})^{\beta} )\right),
    \label{eq:sm}
\end{dmath}
where $n$ is the number of observations. We decide not to include beta in the minimisation because, as it is strongly dependent on the shape of the end effector, once a good value is found, the benefits of fine tuning it would not compensate for the increase in the complexity of the minimisation.
Repeating the dynamic test at different points on the body, it is
possible to create a 3D map of the inspected surface. The 3D map is
then smoothly interpolated using the \texttt{GRIDFIT}
library~\cite{d2005surface}. 
This geometric representation of the surface is then augmented with
elasticity and viscosity information reconstructed using Gaussian
Process Regression
(GPR)~\cite{williams2006gaussian,Chalasani2016ConcurrentPalpation}. In
its standard form, a GPR predicts a scalar value given a (possibly)
multidimensional input. Therefore, we fit two GPRs that predict
elasticity and viscosity given the indenter position. GPR parameters
are learned from the collected data where the input is the indenter
position and the output is the elasticity or the
viscosity. 


\subsection{Variable Impedance Control}\label{ssec:optimization}
The closed-loop behaviour of a Cartesian compliant controller relates the external wrenches applied to the end-effector of the robot to a mass-spring-damper model as follows
\begin{equation}
\boldsymbol{F}^{ext} = \boldsymbol{\Lambda}^d\ddot{\Tilde{\boldsymbol{x}}} + \boldsymbol{D}^d\dot{\Tilde{\boldsymbol{x}}} + \boldsymbol{K}^d\Tilde{\boldsymbol{x}},
    \label{eq:cartesian_impedance_inertia_shaping}
\end{equation}
where $\Tilde{\boldsymbol{x}} = \boldsymbol{x} - \boldsymbol{x}_d \in\mathbb{R}^{m}$ is the Cartesian error computed with respect to the desired Cartesian end-effector pose $\boldsymbol{x}_d$, $\boldsymbol{F}^{ext}\in\mathbb{R}^{m}$ is the external wrench applied on the end-effector, and $\boldsymbol{\Lambda}^d, \boldsymbol{D}^d ,\boldsymbol{K}^d\in\mathbb{R}^{m\times m}$ are the desired Cartesian inertia, damping, and stiffness, respectively. 

Since the stability of the system might be violated in the presence of a variable impedance controller, we enforce the passivity of the system through the concept of passivity of the power port $\dot{\boldsymbol{x}}^T\boldsymbol{F}^{ext}$.
To do so, we introduce the formalism of port-Hamiltonian systems to describe the interaction model of the variable Cartesian impedance augmented with an energy tank~\cite{ferraguti2015energy} with dynamics:
\begin{equation}
    \dot{\textrm{x}}_t = \frac{\sigma}{\textrm{x}_t}\dot{\Tilde{\boldsymbol{x}}}^T\boldsymbol{D}^d\dot{\Tilde{\boldsymbol{x}}} - \frac{\boldsymbol{w}^T}{\textrm{x}_t}\dot{\Tilde{\boldsymbol{x}}},
\end{equation}
where $\textrm{x}_t \in \mathbb{R}$ is the state of the tank, $\sigma \in \{0,1\}$ modulates the energy storage, and $\boldsymbol{w}$ an the extra input of the port-Hamiltonian system defined as:
\begin{equation} \label{eq:variable_stiffness}
    \boldsymbol{w}(t) = \begin{cases} -\boldsymbol{K}^v(t)\Tilde{\boldsymbol{x}} & \mbox{if $T(\textrm{x}_t)>T_{min}$}  \\ 0 & \mbox{otherwise,} \end{cases}
\end{equation}
where $\boldsymbol{K}^v(t)$ is time-varying component of the stiffness
($\boldsymbol{K}^d(t) = \boldsymbol{K}^{min} + \boldsymbol{K}^v(t)$).
At each instant of time, the tank energy is defined by
$T(\textrm{x}_t)=\frac{1}{2}\textrm{x}_t^2$ and
$T_{min}\in\mathbb{R}^+$ is the minimum energy that the tank is
allowed to store. Thanks to~\eqref{eq:variable_stiffness}, we can
infer the condition $T(\textrm{x}_t)>T_{min}$ when the stiffness is
allowed to raise without violating the passivity constraint. However,
this bound does not prevent the energy of the tank to be drained
instantaneously, situation which leads to the complete loss of
performance. For this reason, it is reasonable to further constrain
the power flow of the tank when the energy is extracted from the tank
($\dot{T}(\textrm{x}_t) > \eta \nonumber$), where
$\eta\in\mathbb{R}^-$ is the maximum allowed power.

The viscoelastic body map computed in~\autoref{ssec:parameters_estimation} and the aforementioned passive analysis are then used in a QP that allows online modulation of the stiffness of the Cartesian impedance in~\eqref{eq:cartesian_impedance_inertia_shaping}. The optimisation problem involves a trade-off between the precise tracking of a desired wrench and the necessity to uphold a limited level of stiffness.
Inspired by~\cite{zhao2022hybrid}, we formulated the QP as follows:
\begin{align}
    \min_{\substack{\boldsymbol{K}^d \in \mathbb{R}^{m \times m} }} \: & \frac{1}{2} \left( \| \boldsymbol{F}^{ext} - \boldsymbol{F}^{d} \|_{\boldsymbol{Q}}^2 + \| \boldsymbol{K}^{d} - \boldsymbol{K}^{min} \|_{\boldsymbol{R}}^2 \right) \nonumber \\ 
    \quad  \text{s.t.   } \boldsymbol{K}^{min} &\preccurlyeq \boldsymbol{K}^d \preccurlyeq \boldsymbol{K}^{max} \nonumber \\
    \boldsymbol{F}^{min} &\preccurlyeq \boldsymbol{F}^{ext} \preccurlyeq \boldsymbol{F}^{max} \label{eq:general_QP_formulation} \\
    - \Tilde{\boldsymbol{x}}^{T}\boldsymbol{K}^d \dot{\Tilde{\boldsymbol{x}}} &\leq \sigma \dot{\Tilde{\boldsymbol{x}}}^T \boldsymbol{D}^d \dot{\Tilde{\boldsymbol{x}}} - \Tilde{\boldsymbol{x}}^T 
    \boldsymbol{K}_{min} \dot{\Tilde{\boldsymbol{x}}} + \frac{ T_{t-1} - T_{min}}{\Delta t} \nonumber \\
    - \Tilde{\boldsymbol{x}}^{T}\boldsymbol{K}^d \dot{\Tilde{\boldsymbol{x}}} &\leq \sigma \dot{\Tilde{\boldsymbol{x}}}^T \boldsymbol{D}^d \dot{\Tilde{\boldsymbol{x}}} - \Tilde{\boldsymbol{x}}^T 
    \boldsymbol{K}_{min} \dot{\Tilde{\boldsymbol{x}}} - \eta \nonumber
\end{align}
where $\boldsymbol{Q}$ and $\boldsymbol{R}$ $\in \mathbb{R}^{m\times m}$ are diagonal positive definite weighting matrices, $\boldsymbol{K}^d \in \mathbb{R}^{m\times m}$ is the desired stiffness of the Cartesian impedance controller, $\boldsymbol{K}^{min}$ and $\boldsymbol{K}^{max}$ $\in \mathbb{R}^{m\times m}$ are diagonal matrices representing the minimum and maximum allowed stiffness, $\boldsymbol{F}^{ext} \in \mathbb{R}^{m}$ is the wrench of the impedance interaction model, which can be modelled with \eqref{eq:cartesian_impedance_inertia_shaping}, $\boldsymbol{F}^d \in \mathbb{R}^{m}$ is the desired interaction wrench and $\boldsymbol{F}^{max}/\boldsymbol{F}^{min} \in \mathbb{R}^{m}$ is the maximum/minimum wrench that the robot can exert. The symbol $\preccurlyeq$ represents the matrix inequality. 
The last two constraints limit the maximum energy $T - T_{min}$ which can be injected in the system and the rate $|\eta|$ at which the energy is injected, and are obtained from $T \ge T_{min}$ and $\dot{T} \ge \eta$. 
$1/\Delta t$ is the controller frequency. 
When $T < T_{min}$, the stiffness decreases to its minimum ($\boldsymbol{K}^d = \boldsymbol{K}^{min}$). 

Given the formulation of the optimisation problem in~\eqref{eq:general_QP_formulation}, in this paper we propose two strategies to set the desired and the minimum force, $\boldsymbol{F}^{d}$ and $\boldsymbol{F}^{min}$ in the problem according to~\eqref{eq:hunt_cross_ee}. From now on, we will focus on the vertical component of these forces, denoted as $F_{z,d}$ and $F_{z,min}$. The two strategies are defined as follows:
\begin{enumerate}
    \item Variable Stiffness with Constant Force (\textit{VS-CF}): \\ $F_{z,d} = F_{body}^{ref}$ constant and $F_{z,min}(\varepsilon_{max})$,
    \vspace{1mm}
    \item Variable Stiffness with Variable Force (\textit{VS-VF}): \\ $F_{z,d} = F_{body}(\varepsilon_d)$ and $F_{z,min}$ constant.
\end{enumerate}
In VS-CF the objective is to achieve a reference force, similar to what happens in force control. This force is constrained, however, to meet the condition of maximum penetration that the end effector can have in the body. This constraint is expressed in the minimum force that can be generated, i.e., 
\begin{equation}
    F_{z,min}(\varepsilon_{max},x,y) = \kappa_{x,y} \varepsilon_{max}^{\beta} + \lambda_{x,y} \Dot{\varepsilon} \varepsilon_{max}^{\beta},
    \label{eq:f_max_pen}
\end{equation}
where $\kappa_{x,y}$ and $\lambda_{x,y}$ are the values of the HC
model at that point, and $\varepsilon_{max}$ is the maximum
penetration. We assume that the stiffness of the body is constant up to a certain depth, so it is not necessary to reach $\varepsilon_{max}$ at each palpation during the estimation phase, which may not be clinically safe. The penetration velocity $\Dot{\varepsilon}$ can be
rewritten in function of end effector velocity and surface change in
the direction of movement $\boldsymbol{d}$ as
\begin{equation}
    \Dot{\varepsilon} = - \Dot{z}_{ee} - \nabla z(x, y) \cdot \boldsymbol{d}.
\end{equation}
This formulation prevents the robot from sinking excessively into the
softest parts of the body, adding another degree of safety in addition
to those provided by the energy tanks and energy
valves. 
{\em VS-VF} can be seen as the opposite approach to {\em VS-CF}, as it
tries to maintain the desired penetration along the entire trajectory
but avoid crossing the ribs because the force exerted is excessive and
thus may injure the patient. The equation that describes the desired
force is similar to \eqref{eq:f_max_pen}, but instead of using a
$\varepsilon_{max}$ that has to be avoided, $\varepsilon_{d}$ will be
used, that should be kept over all the trajectory.

More details on the QP problem formulation and energy tank constraint definition are available in~\cite{zhao2022hybrid}. However, our approach differs mainly on the desired wrench definition, which was learned from human demonstrations, while here it is defined with the viscoelastic tissue model. Moreover, in the current formulation, the power flow constraint was added.




\section{Experimental Results}
\begin{figure}[t]
    \centering
    \includegraphics[width=0.9\columnwidth]{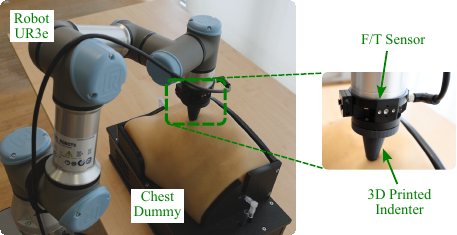}
    \caption{Experimental setup.}
    \label{fig:setup}
\end{figure}

Experiments carried out require the characterisation of the
viscoelasticity map and the repetition of a proof-of-concept
ultrasound in two conditions, namely with and without sudden patient
motion. For each condition, we tested the two proposed variable
impedance strategies (variable stiffness with constant force
\textit{VS-CF} and variable stiffness with variable force
\textit{VS-VF}) and we compared the results with an impedance control
with constant stiffness (\textit{CS}) and a force control with
constant reference (\textit{CF}).

The methodology is evaluated in a proof-of-concept robotic
ultrasonography setup (\autoref{fig:setup}), made of a position controlled manipulator of 6 DoF, the Universal Robot UR3e, a 6-axis
F/T sensor, the SensONE BOTA System mounted on the robot end effector and a 3D printed rigid conical indenter with a spherical
tip ($\SI{2}{\centi\meter}$ radius, $\SI{7}{\centi\meter}$ height),
which is used in parameter estimation and controller
evaluation to simulate the presence of the ultrasound probe. The UR3e
is controlled by means of Forward Dynamics Compliance Control
(FDCC)~\cite{scherzinger2017forward}, which unifies impedance,
admittance, and force control in a single strategy for
position-controlled
robots. 
A similar setup has already been validated with a high-precision measurement device to estimate the elasticity of foams~\cite{beber2024elasticity}.
To prevent potential patient harm and ensure experiment repeatability,
we performed the sonography on a dummy chest, the Blue Phantom
COVID-19 Lung Simulator
($\SI{33}{\centi\meter}\,\times\,\SI{33}{\centi\meter}\,\times\,\SI{23}{\centi\meter}$),
which replicates the 
properties of human tissue, allowing clinicians to practice and
develop the ultrasound imaging skills necessary to diagnose the key
findings consistent, in this specific case, with COVID-19 cases. The
anatomy of the device includes a lung, chest wall, ribs (1-5), and
diaphragm.  The viscoelastic map was computed with MATLAB 2023a, in
particular exploiting \texttt{lsqr} for the least squares method,
\texttt{GRIDFIT}~\cite{d2005surface} for the surface reconstruction,
and \texttt{fitrgp} for the GPR.
The QP problem was instead implemented in C++ using the open-source QPOASES solver~\cite{ferreau2014qpoases} on Ubuntu 22.04. 
All experiments were run on a computer with an AMD Ryzen 9 5900X (24) @ 3.7 GHz $\times$ 12-cores CPU and 32 GB RAM.






\subsection{Viscoelastic Model Validation}
\begin{figure}[t!]
    \centering
    \begin{tikzpicture}
    [spy using outlines={lens={scale=2,rotate=-16}, ellipse, size=2.5cm, height=1.5cm, connect spies},
    ]
        \node[inner sep=0pt] (robot) at (0,0) {\includegraphics{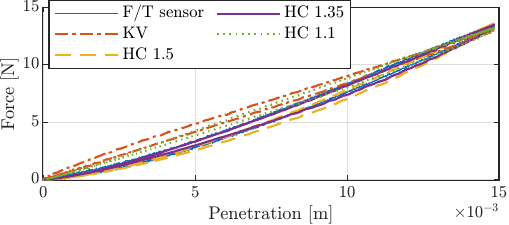}};
    
        \spy [red] on (0.4,0) in node [left,lens={scale=2,rotate=2},rotate=16] at (4,0.2);

    \end{tikzpicture}
    \caption{Load tests obtained with the different models.}
    \label{fig:loadcycle}
\end{figure}
\begin{figure}[t!]
    \centering
    \includegraphics{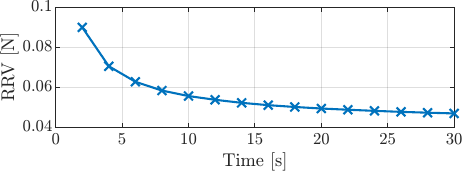}
    \caption{Variation of residual as palpation time changes.}
    \label{fig:residuals}
\end{figure}

The standard procedure to evaluate a viscoelastic model is to perform a load and unload test at constant velocity. This test is done with a Cartesian position controller in the following way: the loading phase is done with constant velocity of $\SI{3}{\centi\meter\cdot\second^{-1}}$ for $\SI{0.75}{\second}$, the load is maintained then for $\SI{10}{\second}$ to reach the equilibrium and finally there is the unloading phase with a velocity of $\SI{1.5}{\centi\meter\cdot\second^{-1}}$ for $\SI{1.5}{\second}$. Results of this test in \autoref{fig:loadcycle} show the limitation of the linear Kelvin-Voight, i.e., for the first phase of the penetration the model prediction is far from the measured force. The Hunt-Crossley model achieves better results in imitating how the body react under a load. 
In \autoref{tab:model_residuals} are shown the residual of the least square \autoref{eq:sm} when varying $\beta$. Minimising the value of this variable over different points of the puppet we obtain that $\beta=1.35$ is the best compromise.
This value of $\beta$ is used to estimate elasticity and viscosity of
the tissue with the sinusoidal palpation and the least square
method. A known issue of the least square method is that is crucial to
have a large amount of observations in order to retrieve precise
estimates. \autoref{fig:residuals} shows how the residuals are
affected by the palpation time; they decrease with exponential behaviour. We choose $\SI{5}{\second}$ as trade-off between length and precision of the estimation.
The palpations on the chest dummy were made at a distance of $\SI{1}{\centi\meter}$ apart so that the surface could be accurately reconstructed. 
\begin{figure}[t!]
    \centering
    \includegraphics[width=\columnwidth]{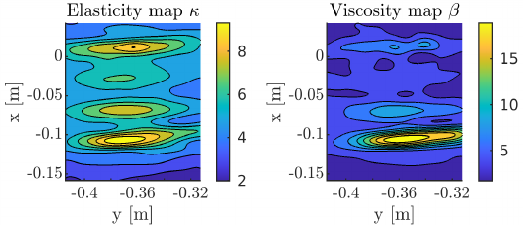}
    \caption{Resulting maps from the GPR computing the values of elasticity and viscosity every $\SI{0.1}{\milli\meter}$.}
    \label{fig:stiffdamp_map}
\end{figure}
\begin{figure}[t!]
    \centering
    \input{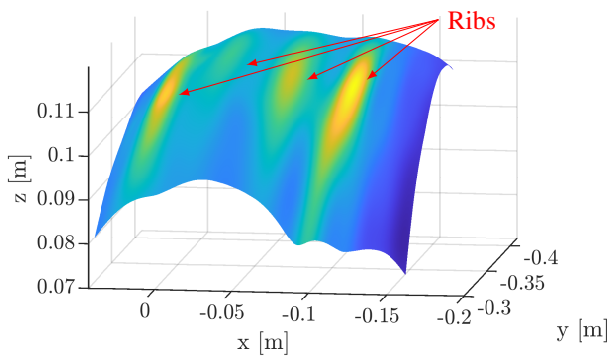}
    \caption{Surface reconstruction with elasticity information. The view is rotated of \SI{180}{\degree} degrees wrt \autoref{fig:stiffdamp_map}  to better show the shape of the body.}
    \label{fig:surface}
\end{figure}

In \autoref{fig:stiffdamp_map} are shown the elasticity and viscosity map computed using the GPR. Focussing on the elasticity map, it can be seen that there are 4 yellow lines, three more evident, and one less; those lines represent the ribs of the chest dummy that are stiffer than the rest of the surface. In \autoref{fig:surface} the dummy-specific elasticity map is projected on the surface reconstructed with the \texttt{GRIDFIT} algorithm. The upward tilt in the bottom left of the figure is the beginning of the shoulder that can be seen in \autoref{fig:setup}. It is interesting to notice that the rib covered by the pectoral muscle seems less stiff than the others covered only by a thin layer of skin.

\begin{table}[t!]
\renewcommand{\arraystretch}{1.2}
\caption{Viscoelastic Model Validation}
\centering
    \begin{tabular}{l c c c}
    \toprule
\multicolumn{4}{c}{\sc{Relative Residual Value [N]}}\\
        \midrule
        Kelvin-Voight& \multicolumn{3}{c}{Hunt-Crossley}\\
        & $\beta$ = 1.5 & $\beta$ = 1.35 & $\beta$ = 1.1 \\ 
        \cline{2-4} 
        0.066 & 0.029 & 0.014 & 0.045 \\
        \bottomrule
    \end{tabular}
\label{tab:model_residuals}
\end{table}

\subsection{Variable Impedance Control Validation}
\begin{figure*}[t]
    \centering
    \includegraphics[trim=0cm 0cm 0cm 0cm,clip,width=\textwidth, height=7.5cm]{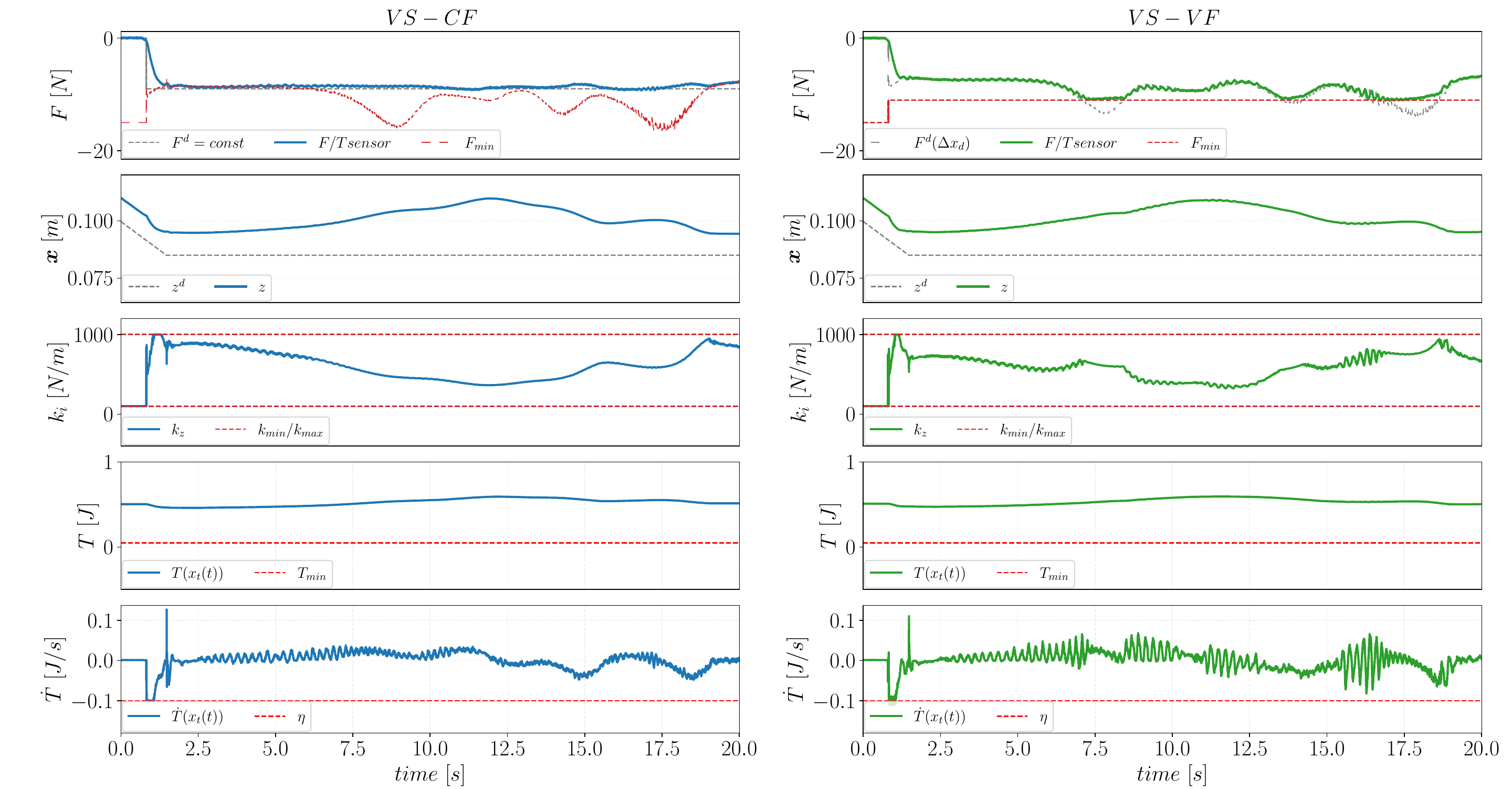}
    \caption{Results of the ultrasound experiment on a dummy chest in nominal conditions. (left) VS-CF, (right) VS-VF. The results of the CS and CF controllers are not reported here for the sake of space. A video of the experiments is available in the multimedia extension and in \url{https://youtu.be/IwQhgzcd4IM}.}
    \label{fig:vs_comparison}
\end{figure*}
\begin{figure*}[!ht]
    \centering
    \includegraphics[trim=0cm 0cm 0cm 0cm,clip,width=\textwidth, height=7.5cm]{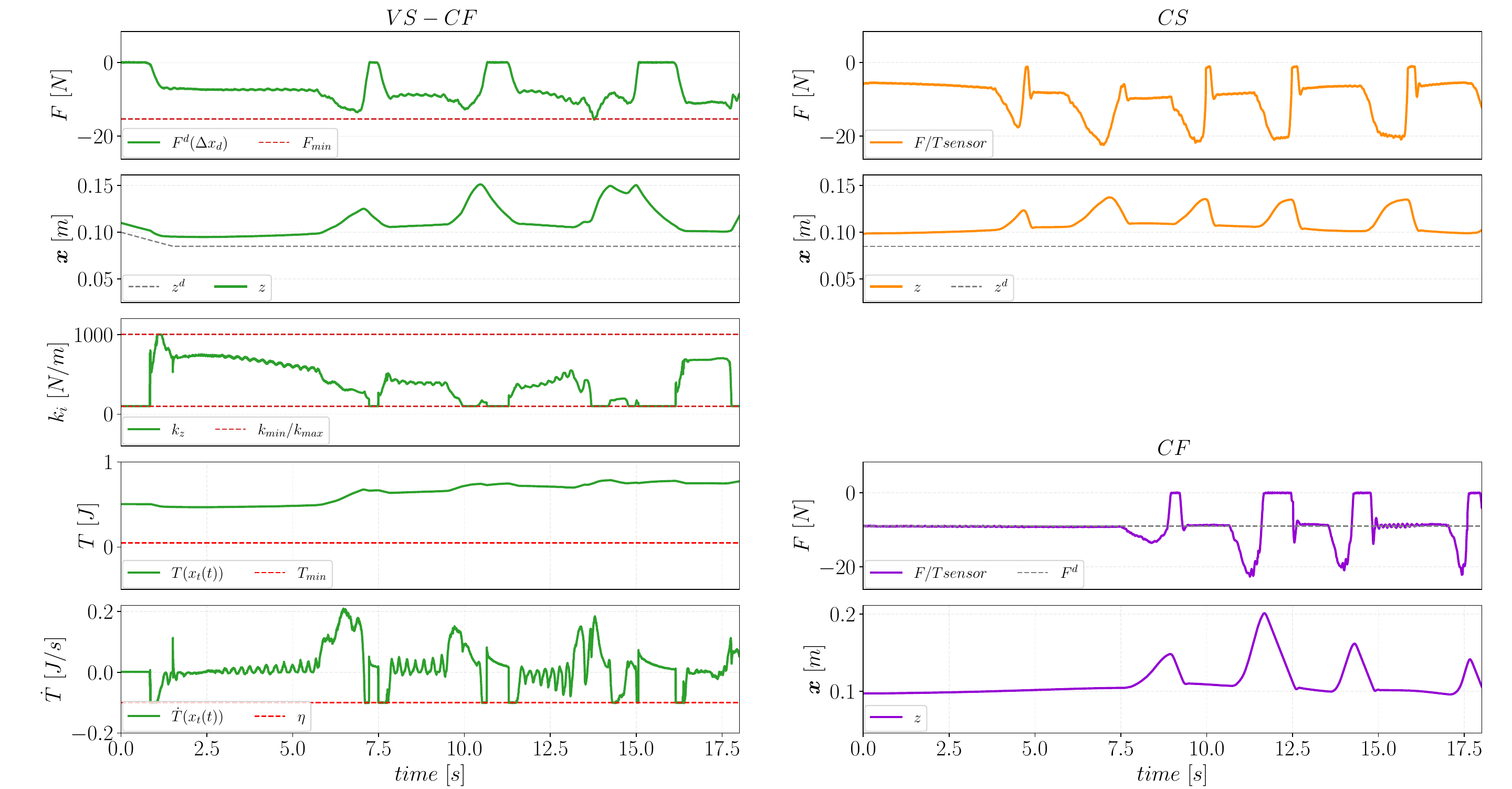}
    \caption{Comparison of ultrasound experiment on a dummy chest with disturbances: (a) VS-VF, (b) CS, (c) CF. }
    \label{fig:comparison_dist}
\end{figure*}

We compare the behaviour of the \textit{CF} and \textit{CS} controller against \textit{VS}-\textit{CF} and \textit{VS}-\textit{VF} that exploit the information of the body
reconstruction. For simplicity, we assume that the value of the rotational stiffness is constant in each direction. The optimisation variable $\boldsymbol{K}^d$ is then a diagonal $3$x$3$ matrix, where each element represents the stiffness in one direction. The minimum and maximum stiffness are identical in every direction and respectively equal to $K_{i,i}^{min}=\SI{100}{N/m}$ and $K_{i,i}^{max}=\SI{1000}{N/m}$.
The inner PD force controller gains, that ensures the force tracking as explained in in~\cite{scherzinger2017forward}, were set to
$k_p = 0.3$ and $k_d = 0.01$ for all the controllers beside the
\textit{CS} where we let the default values of $k_p = 0.05$ and
$k_d = 0.0005$ due to instability problems. 
For experiments, a
reference target is maintained at constant $z$ under the surface of
the dummy to ensure contact and is moved with constant velocity of
$\SI{1}{\centi\meter\cdot\second^{-1}}$ in the longitudinal direction
of the chest dummy. Note that the force controller is hybrid since the
$x$ and $y$ axes are controlled by the compliant controller, while the
$z$ axis is controlled in force, hence it is not possible to give a
reference value to this axis. As expected, the force generated by the
\textit{CS} control is strongly dependent on the shape of the body
(\autoref{fig:comparison_dist}: when the surface is further from
the target, the generated force is stronger than when the surface is
closer. The force controller, instead, can track the target force
without any problem but has the downside that it cannot be controlled
in position, so it is not possible to raise the end-effector from the
surface as in the case of the end of the experiment. Also, in the case of
problems, it would not be possible to stop the end effector except by
resetting the target force it has to reach, which may result in
dangerous motions. The \textit{VS}-\textit{CF} control can track a
reference force as good as a force control
(\autoref{fig:vs_comparison}), but without its downsides. It is
possible to control all the axes when in free motion and also to
suddenly detach the end-effector from the surface in case of
necessity. On top, it uses the information registered by the
reconstruction to cut the reference force when the maximum penetration
is reached preventing the tip from sinking further. The
\textit{VS}-\textit{VF}, instead of tracking a constant force, it is
tracking the force necessary to keep a constant penetration
into the material. In this case, the minimum force avoids pushing too
hard on the stiff points, such as the ribs: at these points,
to achieve the desired penetration, it would require more force than
the maximum force we want to impart on the body.

The same test is done under the action of some disturbances to
demonstrate the inherent safety of this new approach
(\autoref{fig:comparison_dist}). The disturbance consists in raising the dummy to see how the different controls would handle the situation. Given that the
safety of the approaches \textit{VS}-\textit{CF} and
\textit{VS}-\textit{VF} is identical, indeed both use energy tanks and
energy valves to avoid the creation of sudden forces, this test was
done on one of them. We decided to choose the second one given that, in
this approach, the maximum force is limited by a constant. Instead, in
the first approach, the maximum force limit can be subject to
estimation errors leaving room for possible dangerous
behaviour. \autoref{fig:comparison_dist} shows that, as soon as the
end-effector starts to be moved away, the stiffness of the impedance
spring starts decreasing until no force is acting anymore; then, the
spring value is limited to its minimum until the moment of the new
contact. At the moment of contact, the stiffness would suddenly increase, but the presence of the valves partially limits its
growth and restricts the tank from being emptied too quickly.


\section{Conclusion}

This paper presents a novel variable impedance strategy to regulate the interaction forces between the probe and the patient's body in ultrasonography operations. 
This is particularly useful in the case of lung or heart ultrasounds where ribs and muscles are close.
Furthermore, to ensure the stability of the variable impedance, the energy tank was used to ensure the passivity of the system and prevent unsafe behaviour by limiting the minimum energy and the maximum power flow. The experimental results show that the proposed controller outperforms the baseline regarding tracking performance and safety.
However, the approach has some limitations. First, offline viscoelastic estimation requires the patient to remain still for long periods of time, which is not desirable for practical reasons. Thus, an online parameter estimation is paramount~\cite{haddadi2012real}. 
Second, a motion tracking module is required to match the viscoelastic model deformation, as in~\cite{hennersperger2017towards,zhan2020autonomous}. Finally, the control strategy requires a multi-subject evaluation on real patients.

\balance 
\bibliographystyle{IEEEtran}
\bibliography{references_vic,references_med,ref}

\end{document}